\documentclass[letterpaper]{article} % DO NOT CHANGE THIS
\usepackage{aaai2026}  % DO NOT CHANGE THIS
\usepackage{times}  % DO NOT CHANGE THIS
\usepackage{helvet}  % DO NOT CHANGE THIS
\usepackage{courier}  % DO NOT CHANGE THIS
\usepackage[hyphens]{url}  % DO NOT CHANGE THIS
\usepackage{graphicx} % DO NOT CHANGE THIS

\usepackage{booktabs}
\usepackage{xcolor}
\usepackage{enumitem}
\usepackage{amsfonts} 
\usepackage{amsmath}
\usepackage{multirow}

\usepackage{natbib}  % DO NOT CHANGE THIS AND DO NOT ADD ANY OPTIONS TO IT
\usepackage{caption} % DO NOT CHANGE THIS AND DO NOT ADD ANY OPTIONS TO IT
\usepackage{algorithm}
\usepackage{algorithmic}

\usepackage{newfloat}
\usepackage{listings}
\DeclareCaptionStyle{ruled}{labelfont=normalfont,labelsep=colon,strut=off} % DO NOT CHANGE THIS
\floatstyle{ruled}
\newfloat{listing}{tb}{lst}{}
\floatname{listing}{Listing}
\title{Learning to Compress Graphs via Dual Agents for Consistent Topological Robustness Evaluation}
\author {
    Qisen Chai\textsuperscript{\rm 1}, 
    Yansong Wang\textsuperscript{\rm 1}, 
    Junjie Huang\textsuperscript{\rm 1}, 
    Tao Jia\textsuperscript{\rm 1,2,}\thanks{Corresponding Author.}
}
\affiliations {
    \textsuperscript{\rm 1}College of Computer and Information Science, Southwest University\\
    \textsuperscript{\rm 2}College of Computer and Information Science, Chongqing Normal University\\
    cqsllyt@email.swu.edu.cn, yansong0682@email.swu.edu.cn, junjiehuang.cs@outlook.com, tjia@swu.edu.cn
}
\usepackage{bibentry}
\begin{document}

\maketitle

\begin{abstract}
As graph-structured data grow increasingly large, evaluating their robustness under adversarial attacks becomes computationally expensive and difficult to scale. To address this challenge, we propose to compress graphs into compact representations that preserve both topological structure and robustness profile, enabling efficient and reliable evaluation. We propose Cutter, a dual-agent reinforcement learning framework composed of a Vital Detection Agent (VDA) and a Redundancy Detection Agent (RDA), which collaboratively identify structurally vital and redundant nodes for guided compression. Cutter incorporates three key strategies to enhance learning efficiency and compression quality: trajectory-level reward shaping to transform sparse trajectory returns into dense, policy-equivalent learning signals; prototype-based shaping to guide decisions using behavioral patterns from both high- and low-return trajectories; and cross-agent imitation to enable safer and more transferable exploration. Experiments on multiple real-world graphs demonstrate that Cutter generates compressed graphs that retain essential static topological properties and exhibit robustness degradation trends highly consistent with the original graphs under various attack scenarios, thereby significantly improving evaluation efficiency without compromising assessment fidelity.
\end{abstract}

\begin{links}
    \link{Code}{https://github.com/Qisenne/Cutter}
    % \link{Datasets}{https://aaai.org/example/datasets}
    % \link{Extended version}{https://arxiv.org/abs/}
\end{links}

\section{Introduction}

Graphs provide a powerful abstraction for modeling complex systems in the real world, including urban infrastructure, biomolecular interactions, and social networks\cite{battiston2021physics}. These systems often exhibit intricate structures and interdependencies that can be effectively captured using graph-based representations. Among the key properties of such systems, robustness, which refers to the ability to maintain topological integrity under structural perturbations, plays a critical role in ensuring reliable performance\cite{artime2024robustness}. However, as real-world systems grow in size and complexity, their graph representations become increasingly large and computationally expensive to analyze. Most existing robustness evaluation methods rely on topology structural metrics, which are often costly to compute and do not scale well as graph size increases\cite{besta2018survey,zhang2023quantifying}. %This poses a major bottleneck for applying robustness analysis in practical, large-scale settings.

To address this challenge, various graphs reduction methods have been proposed to enable scalable analysis of large-scale graph. These include graph compression~\cite{liu2018graph}, sparsification~\cite{wickman2022generic,wu2020graph}, and sampling~\cite{zhao2020preserving,chen2022preserving}, which aim to preserve critical structural features such as dense subgraphs, semantic backbones, and path-based relevance~\cite{fan2021making,kang2022personalized}. By producing more compact graph representations, these methods help alleviate computational overhead and facilitate downstream tasks. However, they largely ignore whether the compressed graph preserves the robustness profile of the original, that is, how its connectivity degrades under targeted attacks, thereby limiting their utility in robustness analysis scenarios.

% Many systems depend on a small set of structurally critical nodes—such as hubs or articulation points—to sustain global connectivity. Removing these nodes often leads to abrupt and irreversible degradation. Crucially, a node’s robustness contribution depends not only on its own role, but also on the structural support of its neighbors. Some neighbors reinforce the node’s influence, while others are functionally redundant. We leverage this observation by distinguishing supporting from redundant nodes and propose a compression strategy that retains vital nodes and their structural support while removing redundancy. By doing so, we can produce compact graphs that remain aligned with the original in terms of robustness degradation trends under adversarial perturbations.

% Many real-world networks rely on a small set of structurally critical nodes, such as hubs or articulation points, to sustain global connectivity. Removing these nodes can cause abrupt and irreversible collapse. The robustness contribution of a node depends not only on its own role but also on the structural support of its neighbors, some of which are essential while others are redundant. Based on this observation, we distinguish critical and redundant nodes, and define graph compression as reducing graph size through node removal rather than general sparsification or coarsening. This formulation allows us to preserve vital structures while eliminating redundancy, producing compact graphs that remain consistent with the original robustness degradation patterns under adversarial perturbations.

To address this gap, we recognize that certain nodes are structurally critical for maintaining connectivity, and their removal may lead to abrupt and irreversible collapse. A node’s impact on overall robustness depends not only on its own role but also on the structural support of its neighbors, some of which are essential while others are redundant. Building on this observation, we distinguish vital and redundant nodes, and define graph compression as reducing graph size through node removal rather than general sparsification or coarsening. 

In this paper, we propose \textbf{Cutter}, a dual-agent reinforcement learning framework for structural-robustness-preserving graph compression. Cutter comprises two specialized agents: a \textbf{Vital Detection Agent (VDA)} that identifies crucial nodes to retain, and a \textbf{Redundancy Detection Agent (RDA)} that selects non-essential nodes to remove. The two agents share a graph convolutional encoder~\cite{kipf2016semi} for capturing low-level structural features, followed by task-specific sub-encoders~\cite{li2017learning,zhang2022catastrophic} tailored to their respective objectives. This architecture enables coordinated yet specialized decision-making throughout the compression process.

To enhance the coordination and learning efficiency of both agents during the compression process, we design a reinforcement learning optimization mechanism from three complementary perspectives: \textbf{1) Trajectory-Level Return Reward Shaping}. We propose a global shaping strategy that estimates trajectory-level quality using a neural network and allocates fine-grained, step-wise rewards to each state-action pair. To ensure consistency with the trajectory return, an affine transformation is applied to align the cumulative predicted rewards with the actual return. \textbf{2) State-Action-Level Prototype-Constrained Reward Shaping.} To improve the reward network’s ability to capture behavior-level distinctions, we introduce a prototype-based contrastive module that extracts state-action subsequences from trajectories with significant changes in connectivity, encodes them via a recurrent encoder (GRU)\cite{dey2017gate}, and uses them as supervision signals to guide reward estimation. \textbf{3)  Cross-Agent Active–Follow Exploration.} Given the task asymmetry between VDA and RDA, we design an active-follow strategy in which agents alternate between leader and follower roles during training. The follower replicates the leader’s action sequence and evaluates the trajectory under its own objective, enabling safe exploration, and effective knowledge transfer. In summary, the main contributions of this paper are as follows:
\begin{itemize}
    \item We propose a dual-agent reinforcement learning framework for graph compression that preserves the robustness profile of the original graph by jointly identifying vital and redundant nodes. To the best of our knowledge, this is the first method specifically designed to maintain the robustness profile of the original graph during compression.

    \item We design a reinforcement learning optimization mechanism that integrates dense reward shaping, prototype-guided contrastive discrimination, and active-follow cooperative exploration to enhance policy effectiveness and inter-agent collaboration.

    \item We conduct extensive experiments on real-world graph datasets, demonstrating that our method preserves the robustness degradation patterns of the original graphs with high fidelity, even under substantial compression, while maintaining key topological structures.
\end{itemize}

\section{Methodology}

\begin{figure*}[t]
\centering
\includegraphics[width=1\textwidth]{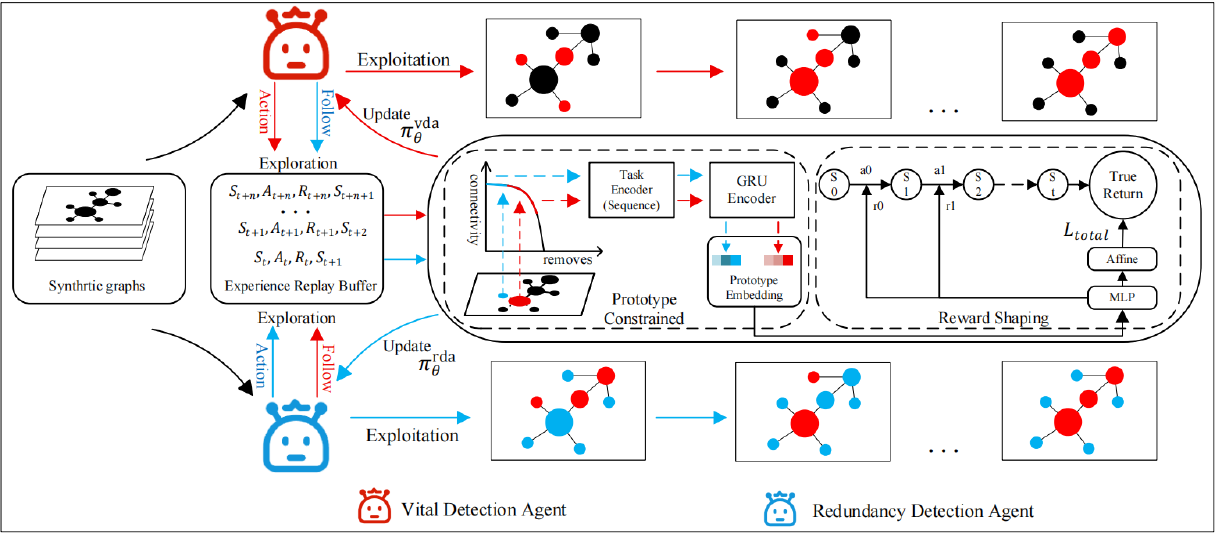} % Reduce the figure size so that it is slightly narrower than the column. Don't use precise values for figure width.This setup will avoid overfull boxes.
\caption{Overview of the Cutter Framework}
\label{fig1}
\end{figure*}

We begin this section by outlining the overall design of our approach. We first define the graph compression task under robustness constraints, then present the architecture of our dual-agent reinforcement learning framework. Finally, we introduce three key modules that drive effective learning and structural-robustness-preserving compression. An overview of the Cutter framework is illustrated in Figure~\ref{fig1}.

\subsection{Problem Definition}
Formally, let \( G = (V, E) \) be an undirected graph with \( |V| = n \) nodes and \( E \subseteq V \times V \) edges.
The robustness of \( G \) is quantified by its pairwise connectivity:
\begin{equation}
F(G) = \sum_{C \in \mathcal{C}(G)} \frac{|C| (|C| - 1)}{2},
\label{pairwise connectivity}
\end{equation}
where \( \mathcal{C}(G) \) denotes the set of connected components in \( G \), and \( |C| \) is the number of nodes in component \( C \). We use pairwise connectivity to measure robustness, as it reflects the number of node pairs that remain connected. This metric is well-suited for real-world systems where overall functionality depends on sustained node-to-node connectivity.\cite{li2021percolation}.

Let $\mathcal{D} \subseteq V$ be a subset of nodes to be removed, yielding a compressed graph $G' = G \setminus \mathcal{D}$ with compression ratio $\rho = 1-\frac{|\mathcal{D}|}{|V|}$. To evaluate the robustness consistency between $G$ and $G'$, we apply a set of adversarial node selection strategies $X$ to both graphs as perturbation methods that simulate targeted attacks on the graph structure. Let $\tilde{G}_x$ and $\tilde{G}'_x$ denote the graphs resulting from applying the attack strategy $x \in X$ to $G$ and $G'$, respectively.  We aim to minimize the discrepancy in robustness under all attack strategies:
\begin{equation}
\min_{\mathcal{D} \subseteq V} \sum_{x \in X} \left| F(\tilde{G}_x) - F(\tilde{G}'_x) \right|
\quad \text{s.t.} \quad |\mathcal{D}| = (1 - \rho)\,|V|.
\label{eq:minimize-robustness-gap}
\end{equation}

To solve the above optimization problem, we formulate the node removal process as a Markov Decision Process (MDP), defined by the tuple \(\langle \mathcal{S}, \mathcal{A}, \mathcal{P}, \mathcal{R}, \gamma \rangle\). 
Each state \(s_t \in \mathcal{S}\) represents the current residual graph at step \(t\), and the action \(a_t \in \mathcal{A}(s_t)\) denotes the node selected for removal. 
The transition function is defined as:
$
\mathcal{P} : \mathcal{S} \times \mathcal{A} \rightarrow \mathcal{S}
$
which updates the state by removing the selected node.  
The reward function \(\mathcal{R}(s_t, a_t)\)  assigns a scalar reward to each state–action pair, which is estimated by a neural network in our framework.
The discount factor \(\gamma \in [0,1]\) controls the weighting of future rewards.

To evaluate the long-term impact of each action, we define the Q-function as the expected cumulative reward:
\begin{equation}
Q(s_t, a_t) = \mathbb{E}_\pi \left[ \sum_{i=0}^{\infty} \gamma^i \mathcal{R}(s_{t+i}, a_{t+i}) \right].
\label{Q}
\end{equation}

Following the Bellman equation, the optimal Q-function satisfies:
\begin{equation}
Q^*(s_t, a_t) = \mathcal{R}(s_t, a_t) + \gamma \cdot \max_{a'} Q^*(s_{t+1}, a').
\label{Q*}
\end{equation}

We adopt standard Deep Q-Networks (DQN)\cite{mnih2015human} to approximate $Q^*$, using experience replay to store and reuse trajectories. This aligns naturally with our graph-based setting, where both agents benefit from sample-efficient reuse of episodic transitions. Each agent maintains its own Q-network and selects actions using an $\varepsilon$-greedy policy. For example:
\begin{align}
\pi^\text{vda}_\theta(a_t \mid s_t) &=
\begin{cases}
\text{random action}, & \text{with probability } \varepsilon \\
\arg\max_{a} Q^\text{vda}(s_t, a), & \text{otherwise},
\end{cases} \notag \\
\pi^\text{rda}_\theta(a_t \mid s_t) &=
\begin{cases}
\text{random action}, & \text{with probability } \varepsilon \\
\arg\max_{a} Q^\text{rda}(s_t, a), & \text{otherwise}.
\end{cases}
\label{eq:eps-greedy}
\end{align}

In addition to separate Q-networks, the agents also employ their own task-specific reward networks, denoted as $\mathcal{R}^\text{vda}_\phi$ and $\mathcal{R}^\text{rda}_\phi$. Each reward network is a learnable module trained to estimate step-wise rewards aligned with the agent’s objective.

\subsection{Agent-Specific Graph Encoder–Decoder}
To enable collaborative learning and task-specific decision-making for dual agents in the graph compression task, we design a reinforcement learning-aware neural architecture tailored for graph-structured data. The architecture comprises three main components: a shared graph encoder that extracts general structural representations, two task-specific encoders for the VDA and RDA, and agent-specific Q-value decoders for predicting node-wise returns.

Given an undirected graph \( G = (V, E) \), the model takes as input a sparse adjacency matrix \( A \in \mathbb{R}^{N \times N} \) and a node feature matrix \( X \in \mathbb{R}^{N \times d_{\text{in}}} \), where \( N = |V| \) is the number of nodes. Since our focus is on topology-driven compression, we operate on graphs without explicit node features. Accordingly, we use an all-one matrix \( X = \mathbf{1}_{N \times d_{\text{in}}} \) as uniform input.

The shared encoder \( f_{\text{shared}} \) produces initial node embeddings by applying a graph convolutional layer:
\[
H^{(0)} = \sigma(A X W_1), \quad H^{(0)} \in \mathbb{R}^{N \times d},
\]
where \( W_1 \in \mathbb{R}^{d_{\text{in}} \times d} \) is a learnable weight matrix and \( \sigma(\cdot) \) denotes the ReLU activation. A global graph embedding is computed via mean pooling:
\[
z^{(0)} = \mathrm{GraphPool}(H^{(0)}),\quad (H^{(0)}, z^{(0)}) = f_{\text{shared}}(A, X).
\]

Based on the shared representation, each agent applies a task-specific encoder to iteratively refine node and graph embeddings. Let \( f_{\text{vda}} \) and \( f_{\text{rda}} \) denote the encoders for VDA and RDA, respectively. These encoders follow the same architecture but are parameterized independently to support task-specific adaptation.

At each layer \( l \), the encoders update node and graph embeddings as follows:
\begin{equation}
H^{(l+1)} = \sigma\left(\mathrm{Linear}\left([A H^{(l)} W_3 \ \| \ z^{(l)} W_2]\right)\right),
\label{H+1}
\end{equation}
\begin{equation}
z^{(l+1)} = \sigma\left(\mathrm{Linear}\left([\mathrm{GraphPool}(H^{(l+1)}) W_3 \ \| \ z^{(l)} W_2]\right)\right).
\label{L+1}
\end{equation}
This process is repeated for $\ell$ layers, yielding the final task-specific representations:
\begin{align*}
H^{(\ell)}_{\text{vda}} &= f_{\text{vda}}(H^{(0)}, z^{(0)}), &
z^{\text{vda}} &= \mathrm{GraphPool}(H^{(\ell)}_{\text{vda}}), \\
H^{(\ell)}_{\text{rda}} &= f_{\text{rda}}(H^{(0)}, z^{(0)}), &
z^{\text{rda}} &= \mathrm{GraphPool}(H^{(\ell)}_{\text{rda}}).
\end{align*}

Finally, each agent uses an independent Q-value decoder to evaluate the expected return of removing a node. Given a node embedding \( h_i \) and a graph embedding \( z \), the Q-value is computed by a multilayer perceptron:
\begin{equation}
Q_i = \mathrm{MLP}([h_i \ \| \ z]),
\label{Qi}
\end{equation}
where the MLP consists of fully connected layers with ReLU activations and layer normalization. This encoder–decoder architecture allows each agent to make context-aware decisions based on both local node properties and global structural cues.

\subsection{Trajectory-Level Return Reward Shaping}
In reinforcement learning, sparse rewards pose a significant challenge, as agents often receive meaningful feedback only after completing an entire trajectory\cite{eschmann2021reward}. This issue is particularly salient in graph-based node removal tasks, where the importance of each action is only revealed in hindsight through its cumulative effect on graph robustness. To address this, we design a reward shaping mechanism that assigns dense, step-wise rewards to individual state-action pairs, guided by trajectory-level returns.

Specifically, in each exploration episode, the agent interacts with the environment to produce a trajectory $\tau = \{(s_0, a_0),\allowbreak\ (s_1, a_1),\allowbreak\ \dots,\allowbreak\ (s_{T-1}, a_{T-1})\}$ consisting of state-action pairs. The overall quality of this trajectory is measured by a ground-truth return $\mathcal{R}_{\text{true}}(\tau)$, which differs across agents due to their task-specific goals.

\noindent\textbf{VDA True Return.}  
Defined as the relative drop in pairwise connectivity between the current original graph \( G_0 \) and the final graph \( G_T \):
\begin{equation}
\mathcal{R}_{\text{true}}^{\text{vda}}(\tau)= P_{\text{conn}} = \frac{F(G_0) - F(G_T)}{F(G_0)},
\label{eq:vda-return}
\end{equation} 
this reward encourages the VDA to identify structurally critical nodes whose removal results in maximal connectivity degradation.

\noindent\textbf{RDA True Return.}  
Defined as a penalty-based score that subtracts from an initial value of 1.0:
\begin{equation}
\mathcal{R}_{\text{true}}^{\text{rda}}(\tau) = 1 - \left(
\omega_1 \cdot P_{\text{conn}} +
\omega_2 \cdot P_{\text{delete}} +
\omega_3 \cdot P_{\text{embed}}
\right),
\label{eq:rda-return}
\end{equation}
where the weights satisfy \( \omega_1 + \omega_2 + \omega_3 = 1 \). The  remaining two terms are defined as:
\begin{equation}
\begin{aligned}
% P_{\text{conn}}^{\text{rda}} &= \frac{F(G_0) - F(G_T)}{F(G_0)}, \\
P_{\text{delete}} &= \frac{|V_{\text{vit}} \cap V_{\text{rem}}|}{|V_{\text{vit}}|}, \\
P_{\text{embed}} &= 1 - \cos\left( \mathbf{h}_{V_{\text{vit}}},\ \mathbf{h}'_{V_{\text{vit}}} \right).
\end{aligned}
\label{eq:rda-penalties}
\end{equation}
Here, \( V_{\text{vit}} \) denotes the set of vital nodes identified by VDA, \( V_{\text{rem}} \) is the set of nodes removed by RDA, and \( \mathbf{h}_{V_{\text{vit}}}, \mathbf{h}'_{V_{\text{vit}}} \) are the mean embeddings of \( V_{\text{vit}} \) before and after node removal. By incorporating feedback from VDA-identified vital nodes, structural connectivity, and semantic embeddings, this reward encourages RDA to preserve essential components while identifying redundancy from multiple perspectives.

\noindent\textbf{Reward Network Design.}  
To propagate trajectory-level supervision to each decision step, we design agent-specific reward networks $\mathcal{R}^\text{vda}_\phi(s_t, a_t)$ and $\mathcal{R}^\text{rda}_\phi(s_t, a_t)$ that estimate the reward for each state-action pair. These networks are trained independently and conditioned on task-specific objectives. The reward for step $t$, denoted as \( r_t \), is computed as:
\begin{equation}
r_t = \tanh \left( W_2 \cdot \mathrm{ReLU}(W_1 [\mathbf{h}_G \ \| \ \mathbf{h}_a]) \right),
\label{eq:reward-step}
\end{equation}
where \( \mathbf{h}_G \) and \( \mathbf{h}_a \) denote the graph-level and action-specific embeddings of the current state-action pair, respectively, and \( W_1 \), \( W_2 \) are learnable weight matrices. The use of \( \tanh(\cdot) \) enables the reward signal to take both positive and negative values, allowing the network to penalize undesirable actions and encourage behavior-sensitive learning.

\noindent\textbf{Reward Network Optimization.}  
The predicted return of a trajectory is obtained by summing the step-wise rewards produced by the reward network:
\begin{equation}
R_{\text{pred}}(\tau) = \sum_{t=0}^{T-1} r_t 
\label{eq:predicted-return}
\end{equation}
% where $r_t = \mathcal{R}_\phi(s_t, a_t)$ denotes the step-wise reward estimated by the reward network.

During training, completed trajectories and their true returns are stored in a replay buffer. The reward network is optimized by minimizing the mean squared error between predicted and true returns:
\begin{equation}
L_{\text{reward}}(\phi) = \mathbb{E}_{\tau \sim \pi} \left[ \left( \mathcal{R}_{\text{true}}(\tau) - R_{\text{pred}}(\tau) \right)^2 \right].
\label{eq:loss-1}
\end{equation}

\noindent\textbf{Affine Return Alignment.}  
In our setting, the true trajectory returns, whether based solely on pairwise connectivity or on multi-faceted penalties including connectivity are always bounded within \([0, 1]\).
 In contrast, the predicted cumulative rewards from the reward network may fall in a broader and potentially unbounded range, such as $[-1, 1]$. This mismatch in scale and distribution can lead to inconsistent gradients and unstable training, especially when minimizing the gap between predicted and true returns.

To address this issue, we introduce an affine transformation that aligns the predicted returns to the scale of the true returns:
\begin{equation}
\mathcal{R}_{\text{affine}}(x) = \alpha x + \beta.
\label{eq:Affine-Return}
\end{equation}
The affine parameters $\alpha$ and $\beta$ are learned via least-squares regression on $(R_{\text{pred}},\ \mathcal{R}_{\text{true}})$ pairs from the replay buffer and frozen after convergence. This ensures that predicted returns align with the true scale, improving training stability.

The final training objective for the reward network becomes:
\begin{equation}
L_{\text{reward}}(\phi) = \mathbb{E}_{\tau \sim \pi} \left[ \left( \mathcal{R}_{\text{true}}(\tau) - \mathcal{R}_{\text{affine}}(R_{\text{pred}}(\tau)) \right)^2 \right].
\label{eq:loss2}
\end{equation}
This transformation not only improves numerical stability but also guarantees policy consistency. A formal justification is provided in the appendix, showing that such affine shaping does not affect the optimal policy under the original return formulation.

\subsection{State–Action-Level Prototype-Constrained Reward Shaping}

While trajectory-level rewards provide global supervision, they often overlook local decision patterns that drive key structural changes. As nodes are progressively removed, certain critical deletions can trigger abrupt collapses in connectivity—akin to first-order phase transitions~\cite{cao2021percolation}. To capture such fine-grained dynamics, we introduce a prototype-constrained reward shaping mechanism that provides step-wise feedback based on similarity to representative behavior fragments from past trajectories.

During training, each agent continuously collects trajectories into an experience buffer. As the buffer grows, we periodically extract the top-$K$ and bottom-$K$ trajectories based on ground-truth returns $\mathcal{R}_{\text{true}}(\tau)$ to guide prototype construction. For each selected trajectory, we compute the agent-specific step-wise true return and identify the most critical decision point. Specifically, we select the step $k^*$ with the highest return for top-$K$ trajectories and the lowest for bottom-$K$:
\begin{equation}
k^* = 
\begin{cases}
\arg\max_t\ \mathcal{R}_{\text{true}}(s_t, a_t), & \text{if top-}K \\
\arg\min_t\ \mathcal{R}_{\text{true}}(s_t, a_t), & \text{if bottom-}K
\end{cases}
\label{eq:critical-step}
\end{equation}
This ensures that the extracted prototypes capture the most representative local decisions characterizing high- and low-quality behaviors, that they are periodically updated based on the agent’s latest experience.

We then extract a fixed-length context window of $n$ state–action pairs preceding that point:
\begin{equation}
\hat{\tau} = \left\{ (s_i, a_i) \right\}_{i = k^*-n}^{k^*-1}.
\label{eq:tao}
\end{equation}

Each state–action pair $(s_t, a_t)$ is encoded into an embedding $h_t$ using an agent-specific encoder $f_{\text{agent}}$, resulting in a temporal sequence $\{h_1, h_2, \dots, h_n\}$. This sequence is then processed by a GRU to capture its dynamic structure:
\begin{equation} %_{\text{agent}}
h_{\text{proto}} = \mathrm{GRU} \left( \left\{ h_i \right\}_{i = k^* - n}^{k^* - 1} \right),
\label{eq:proto-gru}
\end{equation}
where the final hidden state summarizes the behavior pattern that causes critical structural changes.

Aggregating across $K$ trajectories, we compute the mean embeddings for positive and negative prototypes:
\begin{equation}
h_{\text{pos}} = \frac{1}{K} \sum_{i=1}^{K} h^{(i)}_{\text{proto}}, \quad
h_{\text{neg}} = \frac{1}{K} \sum_{i=1}^{K} \tilde{h}^{(i)}_{\text{proto}}.
\label{eq:poneg}
\end{equation}

These prototypes serve as reference patterns for evaluating decision quality.% Notably, they are updated dynamically during training based on the latest experience buffer, ensuring that the shaping signal remains relevant as the agent’s policy evolves. %Moreover, because subtrajectories are of limited length, GRU-based encoding remains both expressive and stable, without over-compressing the temporal information.

When shaping the reward for a new state–action pair $(s_t, a_t)$, we extract its $n$-step historical context, encode it into an embedding:
\begin{equation} %_{\text{agent}}
h_{(s_t, a_t)} = \mathrm{GRU} \left( \left\{ h_i \right\}_{i = t - n}^{t - 1} \right),
\label{eq:proto-embed}
\end{equation}
and compare it to the prototypes via cosine similarity. The reward target is defined as:
\begin{align}
r_{\text{target}}(s_t, a_t) =\ & \text{clip} \Big( \text{sim}_{\cos}(h_{(s_t,a_t)},\ h_{\text{pos}}) \nonumber \\
& -\ \text{sim}_{\cos}(h_{(s_t,a_t)},\ h_{\text{neg}}),\ -1,\ 1 \Big),
\label{eq:proto-reward}
\end{align}
assigning higher rewards to actions aligned with high-quality prototypes, and lower rewards to those resembling negative prototypes.

To propagate this auxiliary signal, the reward network $\mathcal{R}_\phi(s,a)$ is trained to regress toward both the trajectory-level return and the prototype-derived target:
\begin{equation}
L_{\text{proto}}(\phi) = \mathbb{E}_{(s,a) \sim \pi} \left[ \left( \mathcal{R}_\phi(s,a) - r_{\text{target}}(s,a) \right)^2 \right].
\label{eq:proto-loss}
\end{equation}
The overall reward objective combines both learning signals:
\begin{equation}
L_{\text{total}}(\phi) = L_{\text{reward}}(\phi) + \lambda_{\text{proto}} \cdot L_{\text{proto}}(\phi),
\label{eq:total-reward-loss}
\end{equation}
where $\lambda_{\text{proto}}$ is a hyperparameter controlling the influence of local prototype alignment. This hybrid shaping framework allows the agent to learn not only from final outcomes, but also from recurring patterns that characterize positive or negative local decisions.

\subsection{Cross-Agent Active–Follow Exploration}

To encourage cooperation and improve policy generalization, we design a cross-agent exploration strategy in which the VDA and RDA alternate between \emph{active} and \emph{follower} roles. In each episode, both agents operate on the same graph, enabling bidirectional interaction under distinct task objectives.

\noindent\textbf{Phase I: VDA Leads, RDA Follows.}  
VDA explores the graph by executing its policy $\pi^\text{vda}_\theta$, generating a trajectory:
\begin{equation}
\tau^\text{vda}_\text{lead} = \left\{ (s_t, a_t) \right\}_{t=0}^{T}, \quad a_t \sim \pi^\text{vda}_\theta(a \mid s_t),
\label{eq:vda-lead}
\end{equation}
which is evaluated under its ground-truth return $\mathcal{R}^\text{vda}_\text{true}(\tau)$.

 To construct a structural prior for its counterpart, VDA selects a small subset of nodes from its trajectory that are deemed most informative for maintaining structural robustness. Following~\cite{dorogovtsev2008critical}, we retain the top $15\%$ to form an importance set:
\begin{equation}
\mathcal{I}^\text{vda} = \text{Top}{15\%}\left( Q^\text{vda}(s_t, a_t) \mid (s_t, a_t) \in \tau^\text{vda}_\text{lead} \right).
\label{eq:importance-set}
\end{equation}
RDA then replays the same action sequence on the original graph:
\begin{equation}
\tau^\text{rda}_\text{follow} = \left\{ (s'_t, a_t) \right\}_{t=0}^{T},
\label{eq:rda-follow}
\end{equation}
where states $s'_t$ is encoded through RDA’s own graph encoder. This trajectory is evaluated as $\mathcal{R}^\text{rda}_\text{true}(\tau)$ and stored in RDA’s buffer $\mathcal{D}_\text{rda}$.

\noindent\textbf{Phase II: RDA Leads, VDA Follows.}  
RDA explores the graph using its policy conditioned on $\mathcal{I}^\text{vda}$:
\begin{equation}
\tau^\text{rda}_\text{lead} = \left\{ (s_t, a_t) \right\}_{t=0}^{T}, \quad a_t \sim \pi^\text{rda}_\theta(a \mid s_t, \mathcal{I}^\text{vda}),
\label{eq:rda-lead}
\end{equation}
where the importance set provides a global prior highlighting structurally salient regions. VDA then follows this sequence to form:
\begin{equation}
\tau^\text{vda}_\text{follow} = \left\{ (s''_t, a_t) \right\}_{t=0}^{T},
\label{eq:vda-follow}
\end{equation}
% where states $s''_t$ are encoded through VDA’s own graph encoder. The resulting trajectory $\mathcal{R}^\text{vda}_\text{true}(\tau)$ and stored in $\mathcal{D}_\text{vda}$.
where each state $s''_t$ is encoded by VDA’s own graph encoder. 
The resulting trajectory $\tau^\text{vda}_\text{follow}$ is then evaluated using $\mathcal{R}^\text{vda}_\text{true}(\tau)$
and stored in VDA’s buffer $\mathcal{D}_\text{vda}$.

\noindent\textbf{Mutual Evaluation and Training.}
Each agent maintains its own experience buffer $\mathcal{D}_\text{vda}$ or $\mathcal{D}_\text{rda}$, into which both active and follower trajectories are stored. These samples update the policy $\pi^\text{agent}\theta$, reward model $\mathcal{R}^\text{agent}\phi$, and prototype extractor. Reinterpreting peer actions through self-objectives enables implicit transfer across asymmetric tasks, while contrasting decisions helps penalize unsafe patterns for safer, more generalizable policies.

\section{Experiments}
\subsection{Experimental Setup}
\textbf{Datasets.}
We evaluate on five real-world graphs: Cora, Citeseer, PubMed, Coauthor-Physics, and AirTraffic\cite{bojchevski2017deep,rossi2015network}. All are treated as undirected and used without features or labels to isolate structural compression and robustness.

\noindent\textbf{Tasks and Evaluation Metrics.}
To evaluate whether the compressed graph $G'$ preserves the robustness of the original graph $G$, we simulate node removal attacks based on various centrality and influence measures, and record the degradation in pairwise connectivity over time.We use the \textit{Robustness Preservation Similarity} (RPS) to measure the alignment between the degradation curves of $G$ and $G'$ under the same attack strategy. Let $\mathbf{R}{\text{ori}} = [F_0, F_1, \ldots, F_T]$ and $\mathbf{R}{\text{cmp}} = [F_0', F_1', \ldots, F_T']$ denote the pairwise connectivity sequences over $T$ attack steps. RPS is computed as:

\begin{equation}
\text{RPS} = 1 - \frac{1}{T+1} \sum_{t=0}^{T} \left| F_t - F_t' \right|.
\label{rps1}
\end{equation}

A higher RPS indicates that the compressed graph exhibits a connectivity degradation trend closer to that of the original graph, reflecting stronger robustness preservation.

We consider the following eight node attack strategies\cite{freitas2022graph}:
Degree Centrality, Collective Influence, Eigenvector Centrality, Betweenness Centrality, Closeness Centrality, Percolation Centrality, FINDER\cite{fan2020finding}: a reinforcement learning method that removes nodes with the highest expected reward based on state-action embeddings.NIRM \cite{zhang2022dismantling}: ranks nodes by encoding local and global structure, trained on small synthetic graphs via score propagation.

The final robustness score of a method is reported by averaging the RPS scores across all $X$ attack strategies:
\begin{equation}
\text{RPS}_{\text{mean}} = \frac{1}{|X|} \sum_{x \in X} \text{RPS}(x).
\label{rps2}
\end{equation}

\noindent\textbf{Baselines.}
We compare against five representative methods: SparRL~\cite{wickman2022generic}, DPGS~\cite{zhou2021dpgs}, GEC~\cite{meng2024topology}, MCGS~\cite{chen2022preserving}, and HyperSampling~\cite{zhao2020preserving}.All methods are evaluated under uniform compression ratios, with node features set to constant one-vectors to ensure fair comparison in a structure-only setting.

%\noindent\textbf{Models Hyper-parameter and traning.}
%We train the two agents in Cutter using DQN on 5{,}000 randomly generated Barabási–Albert (BA) graphs with 100–150 nodes. In each episode, a graph is sampled, and the agents—VDA and RDA—alternate between leading (exploration) and following (imitation). The leader uses an $\epsilon$-greedy strategy with $\epsilon$ linearly decaying from 1.0 to 0.05 over 5{,}000 episodes. The target Q-network is updated every 100 episodes, and the prototype memory every 10 episodes.Every 50 episodes, RDA is evaluated on a 1{,}000-node BA graph under degree-, betweenness-, and closeness-based attacks, and the model achieving the best connectivity preservation is retained. Since our goal is global sparsification and robustness preservation, we focus on the RDA agent during evaluation. The auxiliary VDA module serves as a training guide, while the learned RDA policy performs compression independently without external input.

\noindent\textbf{Topological Consistency under Compression}  
We evaluate how well each method preserves the static topological structure of the original graph using the Cora dataset as a representative benchmark. Table~1 reports four discrepancy metrics at $\rho$ = 0.5: average degree difference, clustering coefficient difference, path length within the largest component, and SP-Kernel similarity—collectively reflecting both local and global structural fidelity.

Cutter consistently aligns closest to the original graph across all metrics, achieving the best results in clustering, path length, and SP-Kernel similarity, with only a slight deviation in degree. This suggests that Cutter induces the least structural distortion during compression.

% To complement these findings, we examine pairwise connectivity degradation under eight node removal strategies (Figure~\ref{fig2}). At $\rho$=0.5, Cutter’s curve remains nearly identical to the original graph in early stages and shows only minor deviation thereafter. This affirms Cutter’s ability to preserve not only topological features but also robustness patterns throughout the entire attack process.

\begin{figure}[t]
\centering
\includegraphics[width=0.75 \columnwidth]{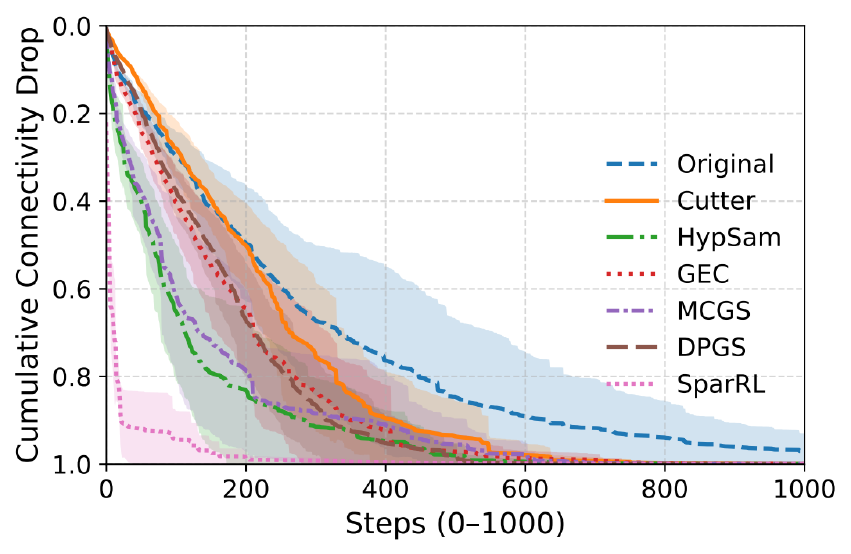} 
\caption{Connectivity drop under various node removal}
\label{fig2}
\end{figure}

To complement these findings, we examine pairwise connectivity degradation under eight node attack strategies (Figure~\ref{fig2}). The shaded regions indicate the variation range across different attack strategies, with lines showing mean trends. At $\rho$ = 0.5, Cutter’s curve remains nearly identical to the original graph in early stages and shows only minor deviation thereafter. This affirms Cutter’s ability to preserve not only topological features but also robustness patterns throughout the entire attack process.

\begin{table}[ht]
\small
\setlength{\tabcolsep}{1.2mm}  % 稍微增宽列距，避免过紧
\renewcommand{\arraystretch}{0.95}
\centering
\label{tab:topo-diff}
\begin{tabular}{lcccc}
\toprule
Method & Degree$\downarrow$ & ClustCoeff$\downarrow$ & PathLen$\downarrow$ & SP-Kernel$\uparrow$ \\
\midrule
SparRL   & 1.309 & 0.097 & 1.856 & 0.871 \\
DPGS     & 0.565 & 0.431 & 1.094 & 0.934 \\
GEC      & 0.161 & 0.163 & 1.045 & 0.960 \\
MCGS     & 0.175 & 0.384 & 0.793 & 0.917 \\
HySamp   & \textbf{0.145} & 0.233 & 1.391 & 0.853 \\
\textbf{Cutter} & 0.167 & \textbf{0.062} & \textbf{0.317} & \textbf{0.972} \\
\bottomrule
\end{tabular}
\caption{Topological differences at $\rho$=0.5. 
$\downarrow$ means lower is better; 
$\uparrow$ means higher is better.}
\end{table}

\noindent\textbf{Robustness Preservation under Adversarial Attacks}
We further assess whether different compression methods can preserve the robustness characteristics of the original graph under adversarial perturbations. The evaluation is conducted on five real-world datasets across three compression ratios: 0.5, 0.3, and 0.1, covering a spectrum from moderate to extreme compression.
We adopt eight predefined node attack strategies, including centrality-based heuristics and learning-based methods. In each case, nodes are ranked by descending importance and progressively removed with a step size of 1\% of the total node count, up to 40\%. Both the original and compressed graphs are attacked using their respectively computed node sequences. At each step, we compute pairwise connectivity to construct degradation curves, and measure the RPS to quantify alignment between the compressed and original graphs.

Table~\ref{tab:rps-singlecol-merged} summarizes the RPS results across all datasets and compression settings. Cutter consistently achieves the highest RPS scores, significantly outperforming all baselines. Notably, even under aggressive compression where only $\rho$=0.1 of the original nodes are retained, Cutter maintains degradation trends that closely mirror those of the original graph, demonstrating its superior ability to preserve structural robustness.
Interestingly, on Citeseer, baseline methods such as GEC and MCGS exhibit robustness similar to Cutter when the compression ratio is moderate. However, their performance deteriorates significantly as the compression becomes more aggressive, particularly at $\rho$=0.1. This observation suggests that robustness preserved under light compression does not necessarily extend to more extreme scenarios. One possible explanation is that these methods rely heavily on specific structural features, such as dense local neighborhoods or community-like clusters, which tend to become unstable or disappear entirely under high compression. In contrast, Cutter maintains stable and consistent robustness across all compression levels, indicating stronger generalizability and adaptability in extreme settings.

\begin{table}[ht]
\small
\setlength{\tabcolsep}{1.2mm}
\renewcommand{\arraystretch}{0.95}
\centering
\begin{tabular}{llccccc}
\toprule
$(\rho)$ & Method & Cora & Cite & PubM & Phys & AirT \\
\midrule
\multirow{7}{*}{0.5}
& SparRL         & 0.822 & 0.927 & 0.938 & 0.826 & 0.837 \\
%& L-Spar         & 0.81 & 0.92 & 0.91 & 0.90 & 0.83 \\
& DPGS           & 0.916 & 0.961 & 0.931 & 0.805 & 0.920 \\
& GEC            & 0.912 & 0.958 & 0.893 & 0.755 & 0.855 \\
& MCGS           & 0.880 & \textbf{0.966} & 0.948 & 0.846 & 0.838 \\
& HySamp         & 0.855 & 0.951 & 0.913 & 0.823 & 0.847 \\
& \textbf{Cutter} & \textbf{0.953} & 0.963 & \textbf{0.965} & \textbf{0.869} & \textbf{0.938} \\
\midrule
\multirow{7}{*}{0.3}
& SparRL         & 0.814 & 0.886 & 0.886 & 0.693 & 0.768 \\
%& L-Spar         & 0.71 & 0.90 & 0.85 & 0.77 & 0.80 \\
& DPGS           & 0.873 & 0.925 & 0.893 & 0.759 & 0.835 \\
& GEC            & 0.851 &\textbf{0.934} & 0.859 & 0.622 & 0.781 \\
& MCGS           & 0.833 & 0.929 & 0.932 & 0.772 & 0.760 \\
& HySamp         & 0.805 & 0.912 & 0.881 & 0.728 & 0.800\\
& \textbf{Cutter} & \textbf{0.885} & 0.920 & \textbf{0.939} & \textbf{0.820} & \textbf{0.875} \\
\midrule
\multirow{7}{*}{0.1}
& SparRL         & 0.781 & 0.853 & 0.829 & 0.512 & 0.724 \\
%& L-Spar         & 0.67 & 0.78 & 0.81 & 0.52 & 0.74 \\
& DPGS           & 0.793 & 0.843 & 0.802 & 0.737 & 0.758 \\
& GEC            & 0.804 & 0.873 & 0.828 & 0.497 & 0.742\\
& MCGS           & 0.774 & 0.881 & 0.861 & 0.690 & 0.746 \\
& HySamp         & 0.777 & 0.878 & 0.854 & 0.624 & 0.717\\
& \textbf{Cutter} & \textbf{0.832} & \textbf{0.895} & \textbf{0.872} & \textbf{0.758} & \textbf{0.780} \\
\bottomrule
\end{tabular}
\caption{RPS scores under different compression ratios. Higher is better.}
\label{tab:rps-singlecol-merged}
\end{table}

\subsection{Ablation Study}

\begin{figure}[t]
\centering
\includegraphics[width=0.75 \columnwidth]{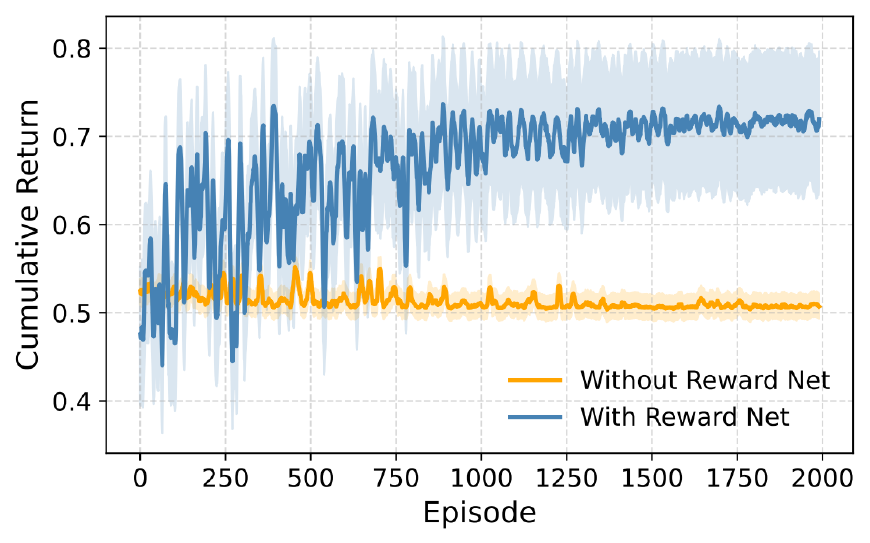} 
\caption{Ablation study on the effect of the reward network}
\label{fig3}
\end{figure}

We evaluate the effect of reward shaping by comparing RDA training with and without the reward network. As shown in Figure~\ref{fig3}, the shaped variant achieves higher average returns (0.668 vs 0.513) and stabilizes over time, while the unshaped agent shows low variance but quickly plateaus at a suboptimal level. Since each episode removes only 30\% of nodes, the unshaped agent operates under strict constraints and struggles to discover effective strategies. The reward network, defined in Equation~\ref{eq:rda-penalties}, combines penalties from connectivity loss, critical-node deletion, and embedding distortion, providing soft guidance that prevents over-compression and preserves structural robustness. 

\section{Related Work}
\textbf{Graph compression} has long been a fundamental technique for reducing the computational burden of graph algorithms. Due to differences in downstream tasks and algorithmic motivations, the literature has developed several closely related yet distinct research directions within this domain. Graph coarsening simplifies the graph structure by merging nodes or edges to improve the efficiency and stability of GNN training. For instance, \cite{chen2021unified} presents a unified coarsening framework, while \cite{meng2024topology} extends the concept of elementary collapse from algebraic topology to guide graph reduction.Graph sampling selects representative nodes or edges to construct a subgraph that preserves essential structural properties. \cite{chen2022preserving} proposes a hybrid sampling approach combining random node sampling and breadth-first search, whereas \cite{zhao2020preserving} adopts a hybrid strategy that integrates triangle-based and cut-point-based heuristics to retain critical minority structures such as super pivots and ties, while balancing majority retention via greedy optimization.
Graph summarization focuses on producing compact representations that capture global structural semantics. \cite{liu2018reducing} introduces CondenSe, an MDL-based framework that identifies diverse patterns (e.g., cliques, stars) through ensemble clustering and assembles supergraphs using k-step selection. \cite{zhou2021dpgs} proposes DPGS, which preserves node degrees and spectral properties through a configuration-based reconstruction scheme and MDL-driven supernode merging. Graph sparsification aims to reduce redundancy by removing or reweighting edges to obtain a sparse approximation of the original graph. Methods in \cite{wickman2022generic} and \cite{wu2020graph} exemplify this direction by preserving structural fidelity while enhancing computational efficiency.

\textbf{Reinforcement learning (RL)} has proven effective for graph-related tasks due to its ability to model sequential decisions and handle delayed feedback. Prior work includes estimating node importance for robustness evaluation~\cite{fan2020finding}, enabling long-range reasoning through dual-agent coordination~\cite{zhang2022learning}, and formulating graph sparsification as an RL problem~\cite{wickman2022generic}. A central challenge in these applications is the sparse reward issue, where agents receive feedback only at the end of long trajectories, limiting learning efficiency. To alleviate this, reward shaping introduces intermediate signals, with potential-based shaping preserving policy optimality~\cite{ng1999policy}. Recent approaches refine shaping via embedding similarity, action confidence, or trajectory-level guidance, including self-supervised ranking preservation~\cite{memarian2021self} and shaping agents modeled as Markov games~\cite{mguni2023learning}. Our work is also related to experience transfer~\cite{brys2015reinforcement} and safe exploration under structural constraints~\cite{karimpanal2020learning}.

\section{Conclusion}
We propose Cutter, a dual-agent reinforcement learning framework for graph compression that preserves structural robustness under adversarial attacks. By integrating reward shaping, prototype-guided learning, and cross-agent exploration, Cutter effectively addresses the sparse reward challenge in graph decision-making. Experiments on real-world graphs show that it achieves high compression while maintaining robustness patterns under targeted node removal. Cutter runs independently at inference time, requiring no external modules or handcrafted rules. Though currently designed for topology-only compression, it can be extended to incorporate attributes and support tasks like GCN training or link prediction, demonstrating strong potential for scalable, robustness-aware graph compression.

\section{Appendix}
\subsection{Policy Invariance under Return-based Reward Assignment}
We first show that using a proxy return function $R_{\phi}(\tau)$ that preserves the trajectory ranking of the true return $R_{\text{true}}(\tau)$ does not alter the optimal policy.

Let $\tau = \{(s_0, a_0), (s_1, a_1), \ldots, (s_{T-1}, a_{T-1})\}$ be a trajectory, and define:

\begin{align}
R_{\text{true}}(\tau) &:= \sum_{t=0}^{T-1} r(s_t, a_t)
\label{eq:true-return-2} \\
R_{\phi}(\tau) &:= \sum_{t=0}^{T-1} R_{\phi}(s_t, a_t)
\label{eq:proxy-return}
\end{align}

Assume that $R_{\phi}$ induces the same **total order** over trajectories as $R_{\text{true}}$, i.e.,
\begin{equation}
R_{\phi}(\tau_1) > R_{\phi}(\tau_2) \iff R_{\text{true}}(\tau_1) > R_{\text{true}}(\tau_2)
\label{eq:order-equiv}
\end{equation}

This implies that the reward functions $R_{\phi}$ and $R_{\text{true}}$ are **order-equivalent**. Following the formalism in~\cite{memarian2021self}, we consider a deterministic MDP $\mathcal{M} = \langle S, A, T, \gamma \rangle$ and define:

\begin{itemize}
\item $\mathcal{T}_{s,a}$: the set of all trajectories starting from $(s,a)$
\item $Q^*_{r}(s,a)$: the optimal action-value under reward function $r$
\item $\tau^*(s,a)$: the trajectory that yields the highest return from $(s,a)$ under $r$
\end{itemize}

Then:
\begin{equation}
Q^*_{r}(s,a) = \max_{\tau \in \mathcal{T}_{s,a}} R_{r}(\tau), \quad \tau^*(s,a) = \arg\max_{\tau \in \mathcal{T}_{s,a}} R_{r}(\tau)
\label{eq:qstar-def}
\end{equation}

A deterministic policy $\pi$ is optimal under reward function $r$ if:
\begin{equation}
\pi^*_r(s) = \arg\max_{a \in A} Q^*_r(s, a)
\label{eq:policy-def}
\end{equation}

Now suppose $R_{\phi} \equiv R_{\text{true}}$ in the total-order sense (Eq.~\ref{eq:order-equiv}). Then for any $s \in S$, $a, b \in A$:
\begin{align}
Q^*_{\text{true}}(s, a) > Q^*_{\text{true}}(s, b)
&\iff R_{\text{true}}(\tau^*(s,a)) > R_{\text{true}}(\tau^*(s,b)) \nonumber \\
&\iff R_{\phi}(\tau^*(s,a)) > R_{\phi}(\tau^*(s,b)) \nonumber \\
&\iff Q^*_{\phi}(s, a) > Q^*_{\phi}(s, b)
\label{eq:equiv-q}
\end{align}

Thus, for all $s$, the maximizing action remains unchanged:
\begin{equation}
\arg\max_{a} Q^*_{\text{true}}(s, a) = \arg\max_{a} Q^*_{\phi}(s, a)
\label{eq:policy-eq}
\end{equation}

Therefore, the set of optimal policies under $R_{\phi}$ is identical to that under $R_{\text{true}}$:
\begin{equation}
\arg\max_{\pi} \mathbb{E}_{\tau \sim \pi} [R_{\text{true}}(\tau)] = \arg\max_{\pi} \mathbb{E}_{\tau \sim \pi} [R_{\phi}(\tau)]
\label{eq:final-policy-equiv}
\end{equation}

This result is consistent with Theorem 1 in ~\cite{memarian2021self}, which states that two reward functions that are total-order equivalent yield identical sets of optimal policies in deterministic MDPs.

\subsection{Policy Invariance under Affine Transformation}

We now consider an affine transformation applied to the proxy return:
\begin{equation}
R_{\text{affine}}(\tau) = \alpha \cdot R_{\phi}(\tau) + \beta
\label{eq:affine-def}
\end{equation}
where $\alpha > 0$ and $\beta$ are constants used to align the value range of $R_{\phi}$ with that of $R_{\text{true}}$.

Let
\begin{align}
[L_{\phi}, U_{\phi}] &= \text{range of } R_{\phi}(\tau) \nonumber \\
[L_{\text{true}}, U_{\text{true}}] &= \text{range of } R_{\text{true}}(\tau)
\end{align}
Then choose $(\alpha, \beta)$ such that:
\begin{equation}
\begin{cases}
\alpha \cdot L_{\phi} + \beta = L_{\text{true}} \\
\alpha \cdot U_{\phi} + \beta = U_{\text{true}}
\end{cases}
\quad \Rightarrow \quad
\begin{cases}
\alpha = \dfrac{U_{\text{true}} - L_{\text{true}}}{U_{\phi} - L_{\phi}} \\[6pt]
\beta = L_{\text{true}} - \alpha \cdot L_{\phi}
\end{cases}
\label{eq:affine-params}
\end{equation}

Since $\alpha > 0$, the transformation is strictly increasing:
\begin{equation}
R_{\phi}(\tau_1) > R_{\phi}(\tau_2) \Rightarrow R_{\text{affine}}(\tau_1) > R_{\text{affine}}(\tau_2)
\label{eq:affine-monotonicity}
\end{equation}

Define the expected return under a policy $\pi$:
\begin{align}
J_{\text{true}}(\pi) &= \mathbb{E}_{\tau \sim \pi}[R_{\text{true}}(\tau)]
\label{eq:jtrue} \\
J_{\phi}(\pi) &= \mathbb{E}_{\tau \sim \pi}[R_{\phi}(\tau)]
\label{eq:jphi} \\
J_{\text{affine}}(\pi) &= \mathbb{E}_{\tau \sim \pi}[R_{\text{affine}}(\tau)] = \alpha \cdot J_{\phi}(\pi) + \beta
\label{eq:jaffine}
\end{align}

The ordering over policies is preserved:
\begin{equation}
J_{\phi}(\pi_1) > J_{\phi}(\pi_2) \iff J_{\text{affine}}(\pi_1) > J_{\text{affine}}(\pi_2)
\label{eq:policy-order}
\end{equation}

Hence, the optimal policy remains unchanged:
\begin{equation}
\arg\max_{\pi} J_{\phi}(\pi) = \arg\max_{\pi} J_{\text{affine}}(\pi)
\label{eq:argmax-eq}
\end{equation}

Combining Equations~\eqref{eq:final-policy-equiv} and~\eqref{eq:argmax-eq}, we conclude:
\begin{equation}
\arg\max_{\pi} \mathbb{E}_{\tau}[R_{\text{true}}(\tau)] = \arg\max_{\pi} \mathbb{E}_{\tau}[R_{\text{affine}}(\tau)]
\label{eq:final-result}
\end{equation}

\section{Acknowledgments}
This work is supported by the Natural Science Foundation of China (No. 62402398, No. 72374173), the Natural Science Foundation of Chongqing (No. CSTB2025NSCQ-GPX1082), the University Innovation Research Group of Chongqing(No.CXQT21005), the Chongqing  Graduate Research and Innovation Project (No. CYB23124), the Fundamental Research Funds for the Central Universities (No. SWU-KR24025, No. SWU-XDJH202303) and the High Performance Computing clusters at Southwest University.

\bibliography{aaai2026}

@article{artime2024robustness,
  title={Robustness and resilience of complex networks},
  author={Artime, Oriol and Grassia, Marco and De Domenico, Manlio and Gleeson, James P and Makse, Hern{\'a}n A and Mangioni, Giuseppe and Perc, Matja{\v{z}} and Radicchi, Filippo},
  journal={Nature Reviews Physics},
  volume={6},
  number={2},
  pages={114--131},
  year={2024},
  publisher={Nature Publishing Group UK London}
}

@article{liu2018graph,
  title={Graph summarization methods and applications: A survey},
  author={Liu, Yike and Safavi, Tara and Dighe, Abhilash and Koutra, Danai},
  journal={ACM computing surveys (CSUR)},
  volume={51},
  number={3},
  pages={1--34},
  year={2018},
  publisher={ACM New York, NY, USA}
}

@inproceedings{wickman2022generic,
  title={A generic graph sparsification framework using deep reinforcement learning},
  author={Wickman, Ryan and Zhang, Xiaofei and Li, Weizi},
  booktitle={2022 IEEE International Conference on Data Mining (ICDM)},
  pages={1221--1226},
  year={2022},
  organization={IEEE}
}

@inproceedings{wu2020graph,
  title={Graph sparsification with generative adversarial network},
  author={Wu, Hang-Yang and Chen, Yi-Ling},
  booktitle={2020 IEEE international conference on data mining (ICDM)},
  pages={1328--1333},
  year={2020},
  organization={IEEE}
}

@article{zhao2020preserving,
  title={Preserving minority structures in graph sampling},
  author={Zhao, Ying and Jiang, Haojin and Chen, Qi'an and Qin, Yaqi and Xie, Huixuan and Wu, Yitao and Liu, Shixia and Zhou, Zhiguang and Xia, Jiazhi and Zhou, Fangfang},
  journal={IEEE Transactions on Visualization and Computer Graphics},
  volume={27},
  number={2},
  pages={1698--1708},
  year={2020},
  publisher={IEEE}
}

@article{chen2022preserving,
  title={Preserving the topological properties of complex networks in network sampling},
  author={Chen, Wen-tao and Zeng, An and Cui, Xiao-hua},
  journal={Chaos: An Interdisciplinary Journal of Nonlinear Science},
  volume={32},
  number={3},
  year={2022},
  publisher={AIP Publishing}
}

@article{kipf2016semi,
  title={Semi-supervised classification with graph convolutional networks},
  author={Kipf, Thomas N and Welling, Max},
  journal={arXiv preprint arXiv:1609.02907},
  year={2016}
}

@inproceedings{dey2017gate,
  title={Gate-variants of gated recurrent unit (GRU) neural networks},
  author={Dey, Rahul and Salem, Fathi M},
  booktitle={2017 IEEE 60th international midwest symposium on circuits and systems (MWSCAS)},
  pages={1597--1600},
  year={2017},
  organization={IEEE}
}

@article{mnih2015human,
  title={Human-level control through deep reinforcement learning},
  author={Mnih, Volodymyr and Kavukcuoglu, Koray and Silver, David and Rusu, Andrei A and Veness, Joel and Bellemare, Marc G and Graves, Alex and Riedmiller, Martin and Fidjeland, Andreas K and Ostrovski, Georg and others},
  journal={nature},
  volume={518},
  number={7540},
  pages={529--533},
  year={2015},
  publisher={Nature Publishing Group}
}

@article{freitas2022graph,
  title={Graph vulnerability and robustness: A survey},
  author={Freitas, Scott and Yang, Diyi and Kumar, Srijan and Tong, Hanghang and Chau, Duen Horng},
  journal={IEEE Transactions on Knowledge and Data Engineering},
  volume={35},
  number={6},
  pages={5915--5934},
  year={2022},
  publisher={IEEE}
}

@article{fan2020finding,
  title={Finding key players in complex networks through deep reinforcement learning},
  author={Fan, Changjun and Zeng, Li and Sun, Yizhou and Liu, Yang-Yu},
  journal={Nature machine intelligence},
  volume={2},
  number={6},
  pages={317--324},
  year={2020},
  publisher={Nature Publishing Group UK London}
}

@inproceedings{zhang2022dismantling,
  title={Dismantling complex networks by a neural model trained from tiny networks},
  author={Zhang, Jiazheng and Wang, Bang},
  booktitle={Proceedings of the 31st ACM International Conference on Information \& Knowledge Management},
  pages={2559--2568},
  year={2022}
}

@article{meng2024topology,
  title={Topology-preserving Graph Coarsening: An Elementary Collapse-based Approach},
  author={Meng, Yuchen and Li, Rong-Hua and Lin, Longlong and Li, Xunkai and Wang, Guoren},
  journal={Proceedings of the VLDB Endowment},
  volume={17},
  number={13},
  pages={4760--4772},
  year={2024},
  publisher={VLDB Endowment}
}

@inproceedings{chen2021unified,
  title={A unified lottery ticket hypothesis for graph neural networks},
  author={Chen, Tianlong and Sui, Yongduo and Chen, Xuxi and Zhang, Aston and Wang, Zhangyang},
  booktitle={International conference on machine learning},
  pages={1695--1706},
  year={2021},
  organization={PMLR}
}

@inproceedings{zhou2021dpgs,
  title={Dpgs: Degree-preserving graph summarization},
  author={Zhou, Houquan and Liu, Shenghua and Lee, Kyuhan and Shin, Kijung and Shen, Huawei and Cheng, Xueqi},
  booktitle={Proceedings of the 2021 SIAM International Conference on Data Mining (SDM)},
  pages={280--288},
  year={2021},
  organization={SIAM}
}

@inproceedings{zhang2022learning,
  title={Learning to walk with dual agents for knowledge graph reasoning},
  author={Zhang, Denghui and Yuan, Zixuan and Liu, Hao and Xiong, Hui and others},
  booktitle={Proceedings of the AAAI Conference on artificial intelligence},
  volume={36},
  number={5},
  pages={5932--5941},
  year={2022}
}

@inproceedings{ng1999policy,
  title={Policy invariance under reward transformations: Theory and application to reward shaping},
  author={Ng, Andrew Y and Harada, Daishi and Russell, Stuart},
  booktitle={Icml},
  volume={99},
  pages={278--287},
  year={1999},
  organization={Citeseer}
}

@inproceedings{memarian2021self,
  title={Self-supervised online reward shaping in sparse-reward environments},
  author={Memarian, Farzan and Goo, Wonjoon and Lioutikov, Rudolf and Niekum, Scott and Topcu, Ufuk},
  booktitle={2021 IEEE/RSJ International Conference on Intelligent Robots and Systems (IROS)},
  pages={2369--2375},
  year={2021},
  organization={IEEE}
}

@inproceedings{mguni2023learning,
  title={Learning to shape rewards using a game of two partners},
  author={Mguni, David and Jafferjee, Taher and Wang, Jianhong and Perez-Nieves, Nicolas and Song, Wenbin and Tong, Feifei and Taylor, Matthew and Yang, Tianpei and Dai, Zipeng and Chen, Hui and others},
  booktitle={Proceedings of the AAAI Conference on Artificial Intelligence},
  volume={37},
  number={10},
  pages={11604--11612},
  year={2023}
}

@inproceedings{brys2015reinforcement,
  title={Reinforcement Learning from Demonstration through Shaping.},
  author={Brys, Tim and Harutyunyan, Anna and Suay, Halit Bener and Chernova, Sonia and Taylor, Matthew E and Now{\'e}, Ann},
  booktitle={IJCAI},
  pages={3352--3358},
  year={2015}
}

@inproceedings{karimpanal2020learning,
  title={Learning transferable domain priors for safe exploration in reinforcement learning},
  author={Karimpanal, Thommen George and Rana, Santu and Gupta, Sunil and Tran, Truyen and Venkatesh, Svetha},
  booktitle={2020 International Joint Conference on Neural Networks (IJCNN)},
  pages={1--10},
  year={2020},
  organization={IEEE}
}

@article{besta2018survey,
  title={Survey and taxonomy of lossless graph compression and space-efficient graph representations. arXiv 2018},
  author={Besta, Maciej and Hoefler, Torsten and Besta, Maciej and Hoefler, Torsten},
  journal={arXiv preprint arXiv:1806.01799},
  year={2018}
}

@article{li2017learning,
  title={Learning without forgetting},
  author={Li, Zhizhong and Hoiem, Derek},
  journal={IEEE transactions on pattern analysis and machine intelligence},
  volume={40},
  number={12},
  pages={2935--2947},
  year={2017},
  publisher={IEEE}
}

@article{zhang2022catastrophic,
  title={Catastrophic interference in reinforcement learning: A solution based on context division and knowledge distillation},
  author={Zhang, Tiantian and Wang, Xueqian and Liang, Bin and Yuan, Bo},
  journal={IEEE Transactions on Neural Networks and Learning Systems},
  volume={34},
  number={12},
  pages={9925--9939},
  year={2022},
  publisher={IEEE}
}

@article{li2021percolation,
  title={Percolation on complex networks: Theory and application},
  author={Li, Ming and Liu, Run-Ran and L{\"u}, Linyuan and Hu, Mao-Bin and Xu, Shuqi and Zhang, Yi-Cheng},
  journal={Physics reports},
  volume={907},
  pages={1--68},
  year={2021},
  publisher={Elsevier}
}

@article{cao2021percolation,
  title={Percolation in multilayer complex networks with connectivity and interdependency topological structures},
  author={Cao, Yan-Yun and Liu, Run-Ran and Jia, Chun-Xiao and Wang, Bing-Hong},
  journal={Communications in Nonlinear Science and Numerical Simulation},
  volume={92},
  pages={105492},
  year={2021},
  publisher={Elsevier}
}

@article{eschmann2021reward,
  title={Reward function design in reinforcement learning},
  author={Eschmann, Jonas},
  journal={Reinforcement learning algorithms: Analysis and Applications},
  pages={25--33},
  year={2021},
  publisher={Springer}
}

@inproceedings{fan2021making,
  title={Making graphs compact by lossless contraction},
  author={Fan, Wenfei and Li, Yuanhao and Liu, Muyang and Lu, Can},
  booktitle={Proceedings of the 2021 International Conference on Management of Data},
  pages={472--484},
  year={2021}
}

@article{liu2018reducing,
  title={Reducing large graphs to small supergraphs: a unified approach},
  author={Liu, Yike and Safavi, Tara and Shah, Neil and Koutra, Danai},
  journal={Social Network Analysis and Mining},
  volume={8},
  number={1},
  pages={17},
  year={2018},
  publisher={Springer}
}

@article{battiston2021physics,
  title={The physics of higher-order interactions in complex systems},
  author={Battiston, Federico and Amico, Enrico and Barrat, Alain and Bianconi, Ginestra and Ferraz de Arruda, Guilherme and Franceschiello, Benedetta and Iacopini, Iacopo and K{\'e}fi, Sonia and Latora, Vito and Moreno, Yamir and others},
  journal={Nature physics},
  volume={17},
  number={10},
  pages={1093--1098},
  year={2021},
  publisher={Nature Publishing Group UK London}
}

@article{dorogovtsev2008critical,
  title={Critical phenomena in complex networks},
  author={Dorogovtsev, Sergey N and Goltsev, Alexander V and Mendes, Jos{\'e} FF},
  journal={Reviews of Modern Physics},
  volume={80},
  number={4},
  pages={1275--1335},
  year={2008},
  publisher={APS}
}

@inproceedings{zhang2023quantifying,
  title={Quantifying node importance over network structural stability},
  author={Zhang, Fan and Linghu, Qingyuan and Xie, Jiadong and Wang, Kai and Lin, Xuemin and Zhang, Wenjie},
  booktitle={Proceedings of the 29th ACM SIGKDD Conference on Knowledge Discovery and Data Mining},
  pages={3217--3228},
  year={2023}
}

@inproceedings{kang2022personalized,
  title={Personalized graph summarization: formulation, scalable algorithms, and applications},
  author={Kang, Shinhwan and Lee, Kyuhan and Shin, Kijung},
  booktitle={2022 IEEE 38th International Conference on Data Engineering (ICDE)},
  pages={2319--2332},
  year={2022},
  organization={IEEE}
}

@article{bojchevski2017deep,
  title={Deep gaussian embedding of graphs: Unsupervised inductive learning via ranking},
  author={Bojchevski, Aleksandar and G{\"u}nnemann, Stephan},
  journal={arXiv preprint arXiv:1707.03815},
  year={2017}
}

@inproceedings{rossi2015network,
  title={The network data repository with interactive graph analytics and visualization},
  author={Rossi, Ryan and Ahmed, Nesreen},
  booktitle={Proceedings of the AAAI conference on artificial intelligence},
  volume={29},
  number={1},
  year={2015}
}

\end{document}